\documentclass[sigconf]{acmart}

\usepackage{subfig}
\usepackage{algorithm}
\usepackage{algpseudocode}

\AtBeginDocument{%
  \providecommand\BibTeX{{%
    \normalfont B\kern-0.5em{\scshape i\kern-0.25em b}\kern-0.8em\TeX}}}

\setcopyright{acmcopyright}
\copyrightyear{2021}
\acmYear{2021}
\acmDOI{}

\acmConference[ACM SIGSPATIAL 2021]{}{}{}
\acmBooktitle{}
\acmPrice{}
\acmISBN{}

\begin{document}

\title{Online Adaptation of Parameters using GRU-based Neural Network with BO for Accurate Driving Model}

\author{Zhanhong Yang}
\email{Zhanhong.yan@ibm.com}
\affiliation{
	\institution{IBM Research}
	\city{Tokyo}
	\country{Japan}
}
\author{Satoshi Masuda}
\email{smasuda@jp.ibm.com}
\affiliation{
	\institution{IBM Research}
	\city{Tokyo}
	\country{Japan}
}
\author{Michiaki Tatsubori}
\email{mich@jp.ibm.com}
\affiliation{
	\institution{IBM Research}
	\city{Tokyo}
	\country{Japan}
}

\begin{abstract}
Testing self-driving cars in different areas requires surrounding cars with accordingly different driving styles such as aggressive or conservative styles. A method of numerically measuring and differentiating human driving styles to create a virtual driver with a certain driving style is in demand. However, most methods for measuring human driving styles require thresholds or labels to classify the driving styles, and some require additional questionnaires for drivers about their driving attitude. These limitations are not suitable for creating a large virtual testing environment. Driving models (DMs) simulate human driving styles. Calibrating a DM makes the simulated driving behavior closer to human-driving behavior, and enable the simulation of human-driving cars. Conventional DM-calibrating methods do not take into account that the parameters in a DM vary while driving. These “fixed” calibrating methods cannot reflect an actual interactive driving scenario. In this paper, we propose a DM-calibration method for measuring human driving styles to reproduce real car-following behavior more accurately. The method includes 1) an objective entropy weight method for measuring and clustering human driving styles, and 2) online adaption of DM parameters based on deep learning by combining Bayesian optimization and a gated recurrent unit neural network. We conducted experiments to evaluate the proposed method, and the results indicate that it can be easily used to measure human driver styles. The experiments also showed that we can calibrate a corresponding DM in a virtual testing environment with up to 26\% more accuracy than with fixed calibration methods.
\end{abstract}



\maketitle

\section{Introduction}
Examples of human driving styles are aggressive, normal, and conservative. These styles vary among countries and areas. It is necessary to create a virtual environment with drivers from different countries and areas to fully test a self-driving car. For example, drivers in Japan have a relatively low accident rate than those in the USA \cite{doi:10.1080/13669877.2018.1517384}.Therefore, testing a self-driving car in the USA requires a more aggressive style than in Japan. The following two problems need to be solved to create such a testing environment:
\begin{enumerate}
\item A driver’s driving style is an abstract concept, it is necessary to develop a method of measuring such styles numerically so that all driver styles in one area can be represented to create a corresponding testing environment. 
\item A driving model (DM) should simulate real human driving behavior.
\end{enumerate}
A car-following model describes the movements of a following vehicle (FV) in response to the actions of the leading vehicle (LV) \cite{ZHU2018425} and is used in a simulation environment such as Simulation of Urban Mobility (SUMO) \cite{SUMO2018}. In a DM, a driver’s behavior is controlled by a series math equations. Setting the parameters in a DM to make the simulated behavior closer to actual human driver behavior is called calibrating a DM. A well-calibrated DM enables the simulation of an actual car, but the parameters of the DM are fixed in the simulation \cite{8317836}. 

\begin{figure}[t!]
\centering
\includegraphics[width=80mm,bb=0 0 413 262]{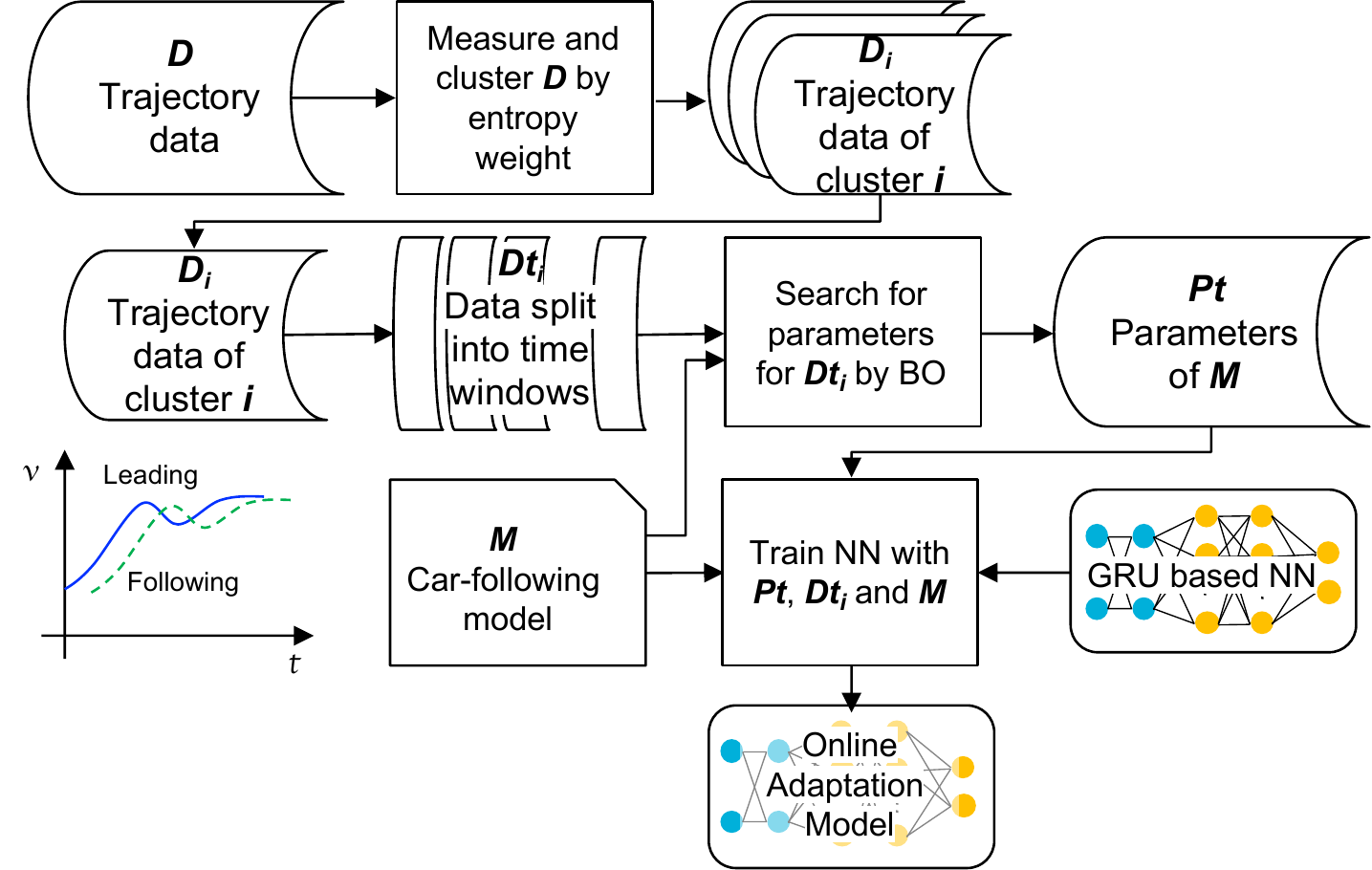}
\caption{Overview of proposed method}
\label{figure:overview}
\end{figure}

We propose a DM-calibration method for measuring real driving styles and for online adaptation of DM parameters using a gated recurrent unit (GRU)-based neural network (NN) to reproduce real car-following behavior more accurately. The proposed method first uses the objective entropy weight method for measuring driving data \cite{8342953} and clustering them. It then splits the driving data by using a short time window (e.g., 0.5 sec) and uses Bayesian optimization (BO) to search for optimized DM parameters for the time windows. It finally trains a GRU-based NN with the optimized DM parameters. We evaluated the proposed method by applying it to the Krauss car-following \cite{7019528} and Wiedemann car-following \cite{wiedemann74} models with the public dataset of driving data Next Generation Simulation (NGSIM), Interstate 80 freeway, collected in California, USA \cite{doi:10.3141/2390-11}. The results indicate that the proposed method is effective in a real simulation environment, i.e., SUMO. The results also indicate that our method can easily numerically differentiate human driver styles throughout a dataset. Therefore, our method could reproduce more accurate DM than conventional DM-calibration methods that do not take into account that the parameters in a DM vary while driving. DM consists of car-following model and lane-changing model in microscopic traffic simulation tool \cite{1504791}. In this paper, we focus on car-following model as a DM. 

In response to the problems and limitations discussed above. Our contributions are as follow:
\begin{enumerate}
\item Our DM-calibration method includes the objective entropy weight method for measuring human driving styles.
\item Our method also involves adaptive calibration.
\item The experimental results indicate that our method can reproduce more accurate DM in simulating a driver’s velocity trajectory than conventional DM-calibration methods.
\end{enumerate}

\section{Related work}
\subsection{Measuring driving styles }
The relationship between driving styles and external (traffic, weather…) and internal conditions (age, gender…) conditions have been investigated. Some studies focused on defining the correct method of labeling driving styles. Researchers classified drivers into discrete classes (aggressive, normal…) or continuous indexing (-1 to 1, for most dangerous to most conservative) Driving-style classification methods are also gaining attention such as rule-based, model-based, and learning-based methods covering supervised or unsupervised learning methods. Most of these methods set absolute criteria and require a clear label for classification \cite{8002632}.

\subsection{Calibrating driving model}
Many models have been developed regarding car-following behavior. Car-following models simulate the longitudinal movement of a vehicle and used to replicate the behavior of a driver when following another vehicle. The Krauss \cite{7019528} and Wiedemann \cite{wiedemann74} models are two such examples. The Krauss model \cite{7019528} can be described as
{\fontsize{8pt}{10pt}\selectfont
\begin{eqnarray}
v_{des} =& \min (v_{max}, v + at_{step}, v_{safe}) \\
v_{safe} =& \nonumber \\
&-b(t_r+\frac{t_i}{2})+\sqrt{b^2(t_r+\frac{t_i}{2})^2+b{v_it_i*\frac{v^2_t}{a}}+2g},
\end{eqnarray}
}
where $v_{des}$ is the desired speed in the simulation environment, $t_r$ is the driver’s overall reaction time, $t_i$ is the driver’s decreasing reaction time; $t_r$ and $t_i$ are the inner parameters defining driving behavior (speed control). To reproduce a driver’s car-following behavior, calibrating $t_r$ and $t_i$ can make $v_{des}$ closer to the real collected data. Therefore, a well-calibrated DM can naturally simulate a human driver’s car-following behavior. The other parameters are maximum deceleration {\it b}, maximum acceleration {\it a}, leading vehicle speed $v_l$, distance (gap) between subject vehicle and leading vehicle $g$, simulation frame interval $t_{step}$, and maximum speed limit $v_{max}$. The forward Euler method is then used to determine vehicle position and speed:
\begin{eqnarray}
V_n(t+\Delta T)&=&V_n(t)+ a_n(t)\cdot \Delta T \\
X_n(t+\Delta T)&=&X_n(t)+ V_n(t)\cdot \Delta T 
\end{eqnarray}

The Wiedemann model \cite{8317836}can be described as
\begin{itemize}
\item $AX$ is the distance to the front vehicle when standing still, and is calculated as $AX(t)$ = $l_{\alpha-1} + AX_{add}$
\item $ABX$ is the desired minimum distance to the front vehicle, and is calculated as $ABX(t)$ = $AX(t) + BX_{add}$.
\item $SDX$ is the maximum distance when following a vehicle, and is calculated as $SDX(t)$ = $SDX_{mult} \cdot ABX(t)$.
\item $SDV$ is the point when the driver notices that he/she is approaching a vehicle with a lower velocity, and is calculated as $SDV(t)$ =$\frac{s_\alpha(t)-AX(t)}{CX}^2$
\item $OPDV$ is the point when the driver notices that the front vehicle is driving away at a higher velocity, and is calculated as $OPDV(t)$ = $SDV (t) \cdot (-OPDV_{add})$.
\end{itemize}
\begin{description}
\item where $AX_{add}$, $BX_{add}$, $SDX_{mult}$, $CX$, and $OPDV_{add}$ are tuning parameters, and $v(t)$ = $min(v_\alpha(t), v_{\alpha-1}(t))$. SUMO simulation uses the 10-parameter version of the Wiedemann model \cite{w99demo}.
\end{description}

A DM-calibration method based on naturalistic driving data has been investigated. However, the dataset is not open, and the inner features are difficult to acquire in large-scale testing \cite{ZHU2018425}. This previous study did not take into account an adaptive calibrating method, and the DM parameters in the simulation remained the same. An adaptive DM-calibration method is necessary to represent a real and complex driving scenario.

\subsection{Bayesian optimization and Gated recurrent unit}
BO is a design strategy for global optimization of black-box functions. It has appeared in machine learning literature as a means of optimizing difficult black box optimizations \cite{10.5555/2999325.2999464}, e.g., BO was used to learn a set of robot parameters that maximize the velocity of a Sony AIBO ERS-7 robot \cite{10.5555/1625275.1625428}. We use BO to search for optimal parameters of a car-following model. The GRU is a gating mechanism in recurrent NNs \cite{cho-etal-2014-learning}, and is used for real-world applications such as predicting traffic flow \cite{7804912}.

\section{Proposed Method}
\subsection{ Measuring driving-style with objective entropy weight}
Our proposed method includes the objective entropy weight method \cite{8342953}. This method is based on the information provided by various attributes to find weights of the information entropy. The steps of calculating the weights are as follows:
\begin{enumerate}
\item For one driving trajectory dataset, construct evaluation matrix \boldmath$E$. Assuming the set includes trajectory data of {\it m} vehicles with ID = 1,2,\ldots,{\it m}, a vehicle with ID = $i$, \boldmath$E$ = $(e^{\prime}_{ij})_{m \times n}$ , where {\it n} = 6, is
{\fontsize{7.5pt}{9pt}\selectfont
\begin{eqnarray}
 E &=& (e^{\prime}_{ij})_{m \times n} \nonumber \\
 &=& \left(
 \begin{array}{ccc}
 {\it Vel_{mean,1}} & {\it Vel_{mean,i}} & {\it Vel_{mean,m}} \\ 
 {\it Vel_{var,1}} & {\it Vel_{var,i}} & {\it Vel_{var,m}} \\
 {\it Acc_{mean,1}} ... & {\it Acc_{mean,i}} ... & {\it Acc_{mean,m}} \\
 {\it Acc_{var,1}} & {\it Acc_{var,i}} & {\it Acc_{var,m}} \\
 {\it H_{s,1}} & {\it H_{s,i}} & {\it H_{s,m}} \\
 {\it H_{t,1}} & {\it H_{t,i}} & {\it H_{t,m}}, \\
 \end{array}
 \right)
\end{eqnarray}
}
where ${\it Vel_{mean}}$ is mean velocity, ${\it Vel_{var}}$ is variance of velocity, ${\it Acc_{mean}}$ is mean acceleration, ${\it Acc_{var}}$ is variance of acceleration, ${\it H_s}$ is average speed headway, and ${\it H_t}$ is average time headway.

\item Normalize \boldmath$E$ into (0,1) and obtain \boldmath$E_{nor}$ = $(e^{\prime}_{ij})_{m \times n}$. For column {\it j}, normalize \boldmath$E$ in accordance with the following equations:

\begin{eqnarray}
e_{ij} =
\begin{cases}
 \frac{Max(e^{\prime}_{ij}) - e^{\prime}_{ij}}{Max(e^{\prime}_{ij}) - Min(e^{\prime}_{ij})} & (e^{\prime}_{ij}<0),\\
 \frac{e^{\prime}_{ij}-Min(e^{\prime}_{ij})}{Max(e^{\prime}_{ij}) - Min(e^{\prime}_{ij})} & (e^{\prime}_{ij}>0).
\end{cases}
\\
Max(e^{\prime}_{ij}) = \max (e^{\prime}_{1j},e^{\prime}_{2j},\ldots e^{\prime}_{mj}) \nonumber \\
Min(e^{\prime}_{ij}) = \max (e^{\prime}_{1j},e^{\prime}_{2j},\ldots e^{\prime}_{mj}) \nonumber
\end{eqnarray}

\item For column {\it j}, calculate the percentage of data weight \boldmath$p_{ij}$:
\begin{eqnarray}
p_{ij} = \frac{e_{ij}}{\sum_{i=i}^m e_{ij}}
\end{eqnarray}

\item Calculate the entropy score \boldmath$ent_{j}$:
\begin{eqnarray}
ent_{j} = -\frac{1}{\ln (m)}\sum_{i=i}^m p_{ij}\ln(p_{ij})
\end{eqnarray}

\item Calculate the entropy weight vector \boldmath$W$:
\begin{eqnarray}
W = w{j} = -\frac{1-ent_{j}}{n-\sum_{i=i}^n ent_{j}}
\end{eqnarray}

\end{enumerate}
For most cases, we consider that one driver drives the same vehicle in the dataset. Then, the exact driving style of vehicle ID={\it i} (also driver ID={\it i}) in \boldmath$E_{nor}$ is measured by a score that is represented as $s_i$:
\begin{eqnarray}
s_i = [e_{i1} \ldots e_{ij} \ldots e_{i6} ]\cdot W
\end{eqnarray}

The objective entropy weight method is used to divide and select the driving trajectory dataset.

\subsection{Online adaptation of DM parameters}
\begin{figure}[t!]
\centering
\includegraphics[width=60mm,bb=0 0 222 165]{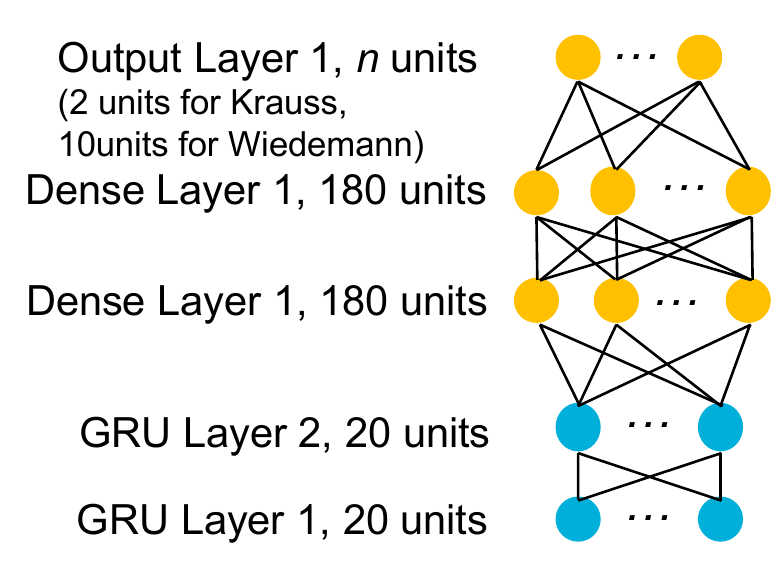}
\caption{Gated recurrent unit (GRU)-based neural-network (NN) structure}
\label{figure:nn_architecture}
\end{figure}

With the proposed method, different driving styles are in one dataset. To create an exact virtual testing environment, however, we need a method for reproducing a driver’s behavior. In simulation, a driver’s behavior is generally controlled using a DM. In this study, we used the Krauss and Wiedemann models as DM models. A previous study proposed many fixed DM-calibration methods \cite{ZHU2018425}. However, the main limitations with these methods are that they do not take into account that the inner parameters can vary when driving (hereafter, these methods are called “fixed calibration methods”). The proposed method uses BO and a GRU-based NN. Figure \ref{figure:nn_architecture} shows the architecture of the GRU-based NN. Assuming that our reproducing target is a subject vehicle with a leading vehicle for a car-following DM. The output should be a policy $\pi$ ($v_p$ , $v_s$) used for the NN, which inputs both the leading vehicle state $v_p$ and subject vehicle state $v_s$, e.g., subject vehicle’s speed trajectory in the past few steps. The proposed method predicts the next possible parameters, which minimizes the difference between the actual and simulated trajectories.
 
Algorithm \ref{algorithm:calibration} shows the steps of the online adaptation of DM parameters. Assuming the target is to reproduce the driving behavior of the subject vehicle by using a DM with $M$ inner parameters sets $pM\{p1, p2,\cdots pM\}$ to be calibrated. The data used for reproduction can be one driver (assuming he/she has an ID=$i$, $s_i$). For driver i in the reproduction dataset, we assume the data from the subject vehicle as $d_{s,i}$, and the data from the pre-ceding as $d_{p,i}$.
 
\renewcommand{\algorithmiccomment}[1]{\bgroup\hfill//~#1\egroup}
\renewcommand{\algorithmicrequire}{\textbf{Input:}}
\renewcommand{\algorithmicensure}{\textbf{Output:}}

\begin{algorithm}[h!]
\caption{Algorithm for online adaption of DM parameters for vehicle $i$}
\label{algorithm:calibration}
\begin{algorithmic}[1]
\Require
$d_{s,i}$, $d_{p,i}$
\Ensure
$\pi$ ($v_p$ , $v_s$)
\Repeat
\State $t$=0, $a$, 2$a$\ldots
\State Split the $d_{s,i}$, $d_{p,i}$ into short time windows every $a$ steps, each window length is denoted as $L$, obtain a sliced dataset from $d_{s,i}$, $d_{p,i}$ as $d^{k=0, L}_p$, $d^{k= a, L+a}_p$ \ldots ; $d^{k=0, L}_s$, $d^{k= a, L+a}_s$ \ldots, where $k$ is the time frame. 
\Repeat
\State Loss= RMSE ($d^{k= j, L+j}_s$ , $d^{k= j, L+j}_sims$)
\State For time step $j$, search for the best inner-parameter set $P_{M, j}$ by BO using $d^{k= j, L+j}_p$ , $d^{k= j, L+j}_s$
\Until{all $d^{k=j, L+j}_p$ , $d^{k= j, L+j}_s$ have been found}
\Until{$d_{s,n}$, $d_{p,n}$ has been found}
\State \boldmath$Train$ $\pi$ ($v_p$ , $v_s$) with input $d^{k=j, L+j}_p$ , $d^{k= j, L+j}_s$ and labels: $P_{M, j}$
\end{algorithmic}
\end{algorithm}
 
\section{Experiments}
\subsection{Dataset}
We conducted experiments on the NGSIM, Interstate 80 freeway (I-80) dataset, and collected data between 4:00 p.m. and 4:15 p.m. on April 13, 2005. The study area was approximately 500 meters (1,640 feet) in length. The dataset consists of comma-separated values and the columns include vehicle ID, position (x, y), velocity, acceleration, following vehicle ID, and so on. The data represent moving vehicles every 0.1 second (one frame). To obtain sufficient trajectory data for the experiments, we selected 94 pairs of leading and following vehicles ({\it I-80 Selected Data}) that have over 70 seconds (700 frames) for their driving, from a total of 1475 vehicles in the dataset.

\subsection{Driving-style score}
\begin{figure}[t!]
\centering
\includegraphics[width=80mm,bb=0 0 461 346]{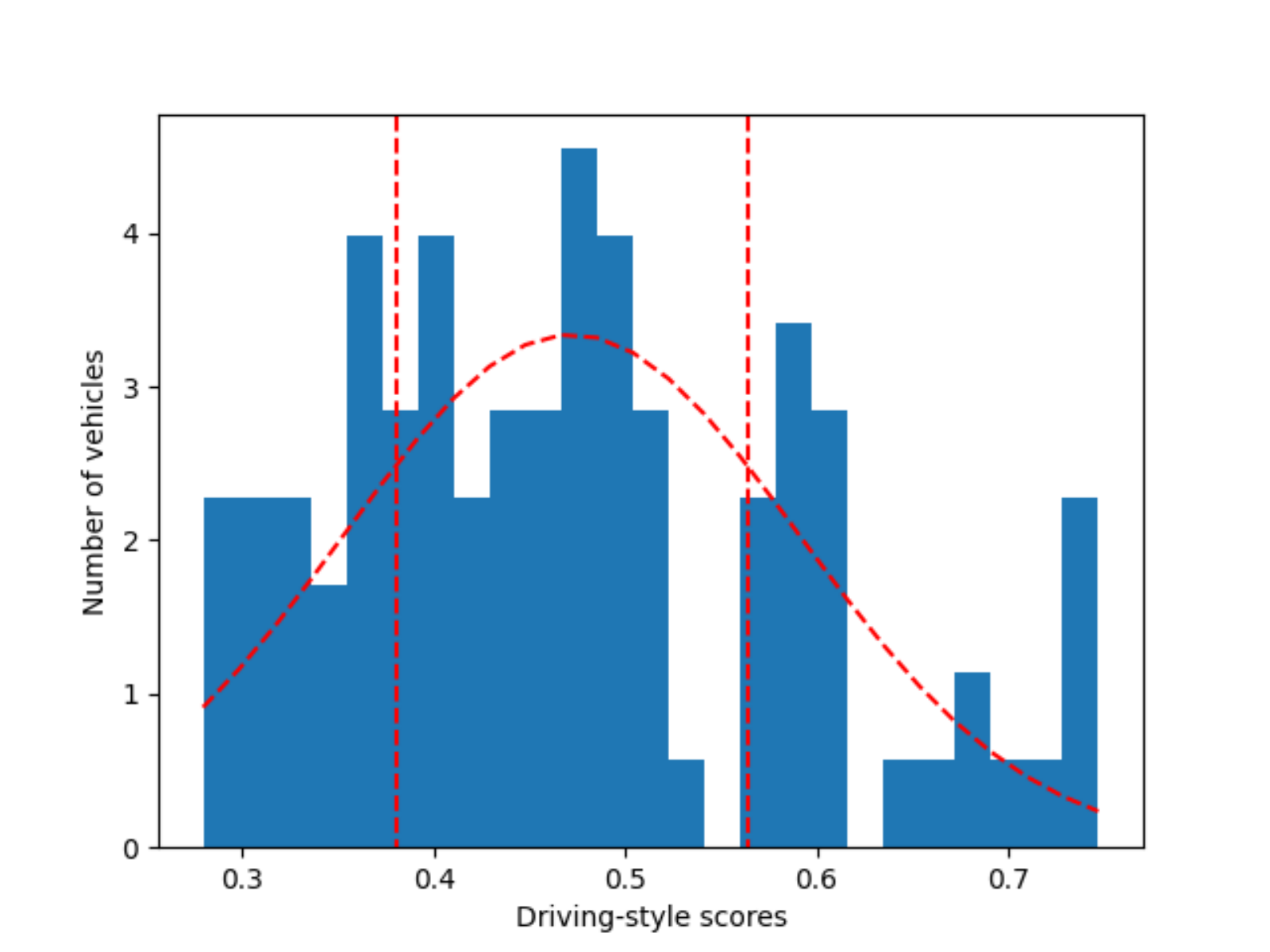}
\caption{Driving-style measurement with distribution of driving styles}
\label{figure:score}
\end{figure}

We calculated driving-style scores $s_i$ for the I-80 Selected Data. Figure \ref{figure:score} shows the driving styles in the distribution. We divided the styles into three clusters on the basis of their percentile; cluster-0 (conservative), which is the bottom 25\% (24 vehicles, driving score $<$0.380), cluster-1 (normal), which is the middle 50\% (46 vehicles, 0.380$\leq$ driving score $\leq$0.564) and cluster-2 (aggressive), which is the top 25\% (24 vehicles, driving score $>$0.564). 

\subsection{Online adaptation}
\subsubsection{Vehicle data for Krauss model}
We first applied the proposed method to each pair of leading and following vehicles in the {\it I-80 Selected Data}. The two car-following models simulated the following vehicle in accordance with the leading vehicle, so we applied the proposed method to the following vehicle. The trajectory data were spliced into 0.5-second (5 frames) short time windows, and the parameters for the sliced dataset were searched for by BO. We used the first 80\% of the trajectory data for training the GRU-based NN with 0.1 for validation split as a hyper-parameter of training the NN. The remaining 20\% of data were selected as the test sets for the GRU-based NN. For example, a pair of leading ID=1 and following ID=11 vehicles has records of 83.9-seconds (839 frames) for its trajectory in the dataset, the data were split into 167 (=839$\div$5) time windows, and the NN training used 134 (=167$\times$80\%) and 33 (=167$\times$20\%) time windows. We used epochs = 500 and batch size = 1 as hyper-parameters of training, as a result of grid search of the hyper-parameters. 

\begin{figure}[t!]
\centering
\includegraphics[width=80mm,bb=0 0 461 346]{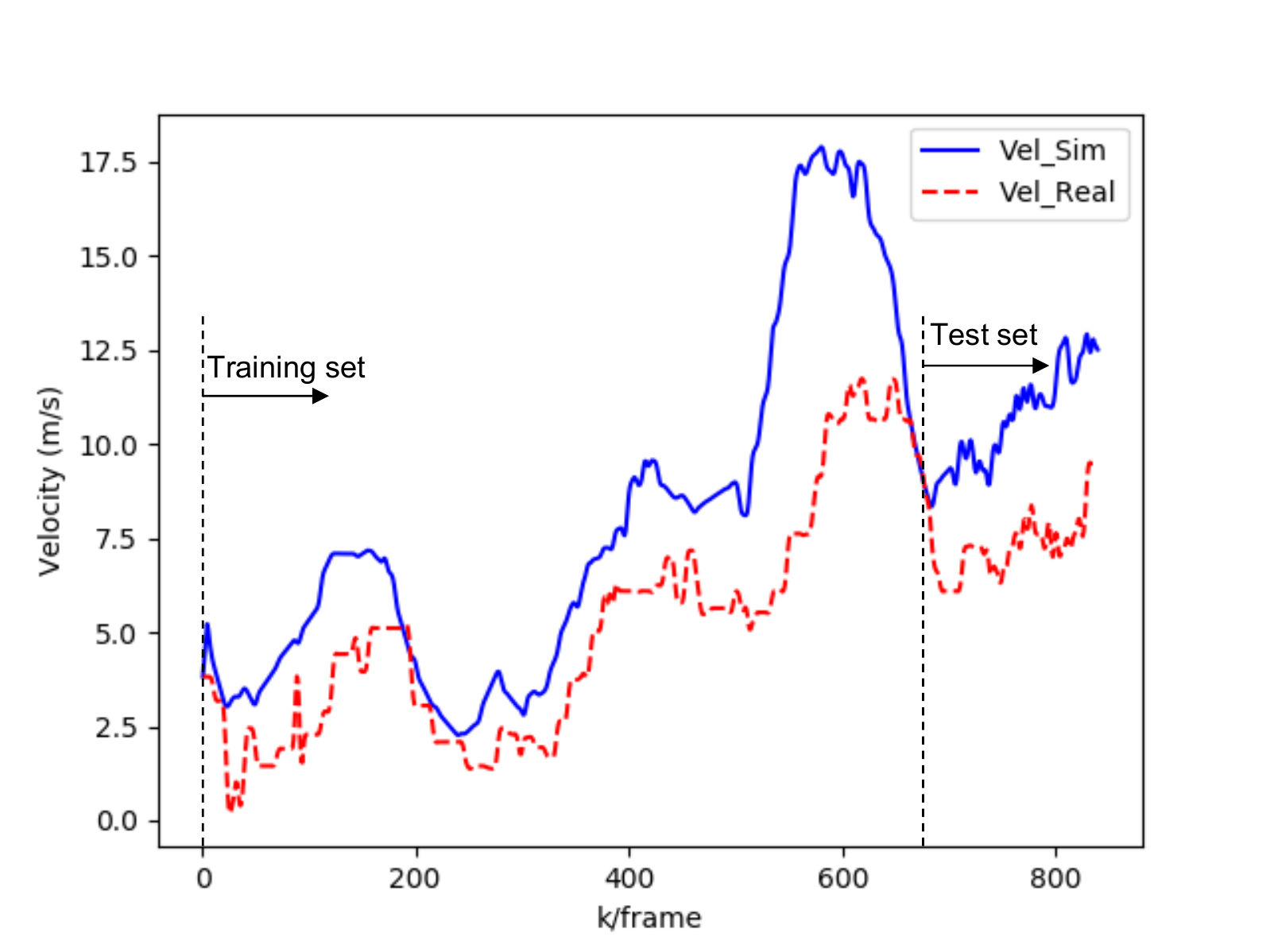}
\caption{Default parameters}
\label{figure:no_calib}
\centering
\includegraphics[width=80mm,bb=0 0 461 346]{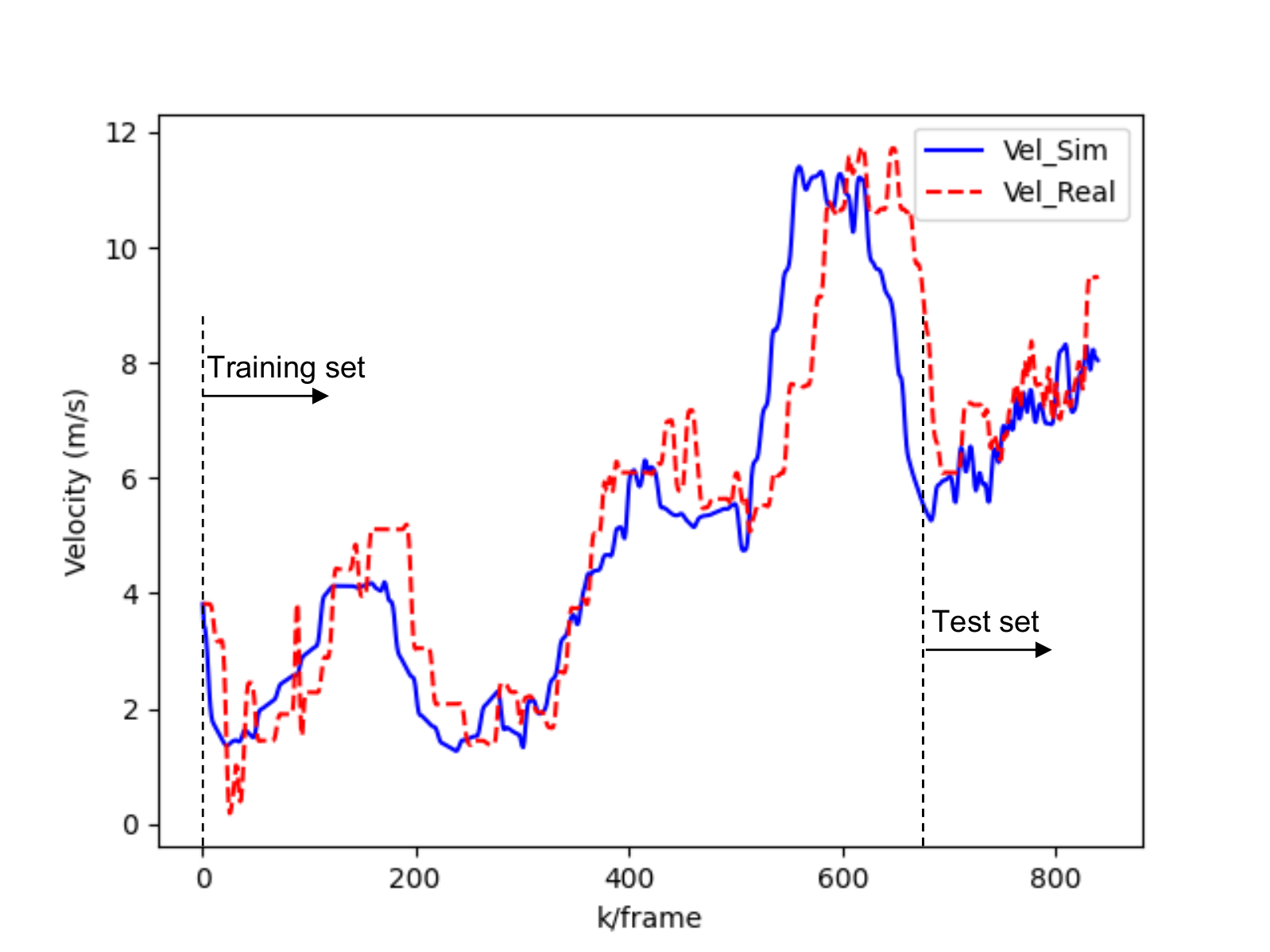}
\caption{Fixed parameters}
\label{figure:bo_calib}
\centering
\includegraphics[width=80mm,bb=0 0 461 346]{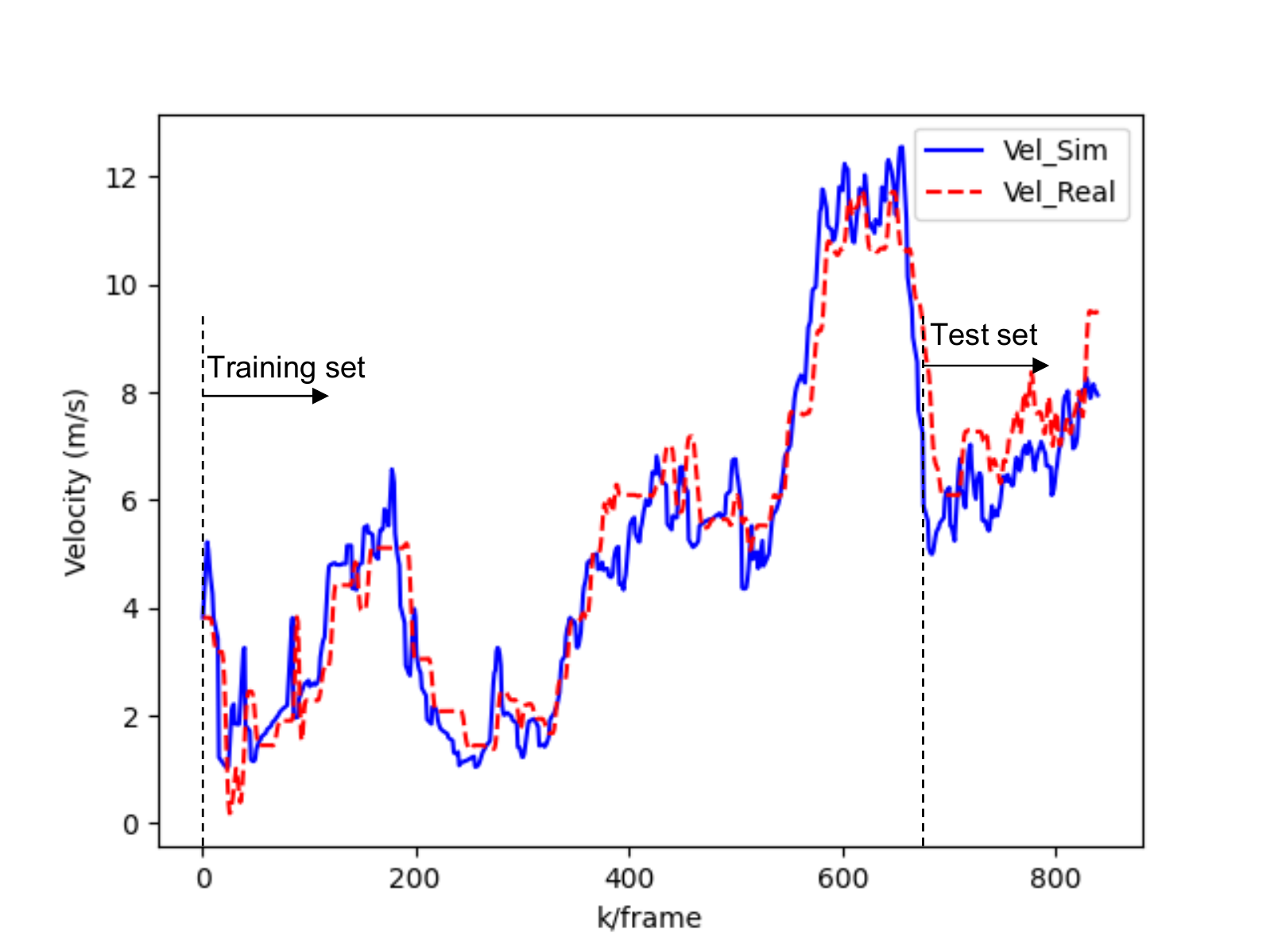}
\caption{Proposed method}
\label{figure:dm_calib}
\end{figure}

We evaluated the proposed method by comparing it with fixed calibration methods; one  using the Krauss model with the initial default parameters, e.g., $t_r$ = 1.5 and $t_i$ = 0.15, (hereafter, the method is called {\it default parameters}) and with {\it fixed} parameters searched for by BO using 80\% of the trajectory data (hereafter, this method is called {\it fixed parameters}). Note that this BO is only used in experiments and different from the BO in the proposed method. Figure \ref{figure:no_calib} shows the results of {\it default parameters} with the velocity of the car-following simulation (Vel\_Sim) and real data (Vel\_Real) for the following vehicle ID=11 trajectory-data frames, Figure \ref{figure:bo_calib} shows the velocity on the same vehicle ID for {\it fixed parameters}, and Figure \ref{figure:dm_calib} shows the results of the proposed method, i.e., online adaptation of parameters. Table \ref{table:krauss-1} shows the root mean square error (RMSE) between Vel\_Sim and Vel\_Real. The proposed method differed from {\it fixed parameters}. We focused on the improvement of the proposed method from {\it fixed parameters} to evaluate the accuracy of the proposed method. In the following vehicle ID=11 case, RMSE of the {\it fixed parameters} was 1.033 and the one of proposed method was 0.666, therefore, RMSE improvement of the proposed method from the {\it fixed parameters} was 35.5\% (=(1.033-0.666)$\div$1.033) (hereafter, this improvement is called {\it RMSE improvement} for evaluation of the proposed method). 

\begin{table}[t!]
\centering
\caption{Experiment on vehicle ID =1 for Krauss model}
\label{table:krauss-1}
\begin{tabular}{p{10mm}|p{11mm}|p{15mm}|p{14mm}||p{13mm}}
\hline 
 & Default param. & Fixed param. (a) & Proposed (b) & Improvement (c) \text{*} \\
\hline 
RMSE & 3.393 & 1.033 & {\bf 0.666} & 35.5\% \\
\hline 
\multicolumn{5}{l}{\text{*} c = (a-b)/a}
\end{tabular}
\end{table}

\subsubsection{Applying Krauss and Wiedemann models to each vehicle pair}
\begin{figure}[t!]
\centering
\includegraphics[width=80mm,bb=0 0 461 346]{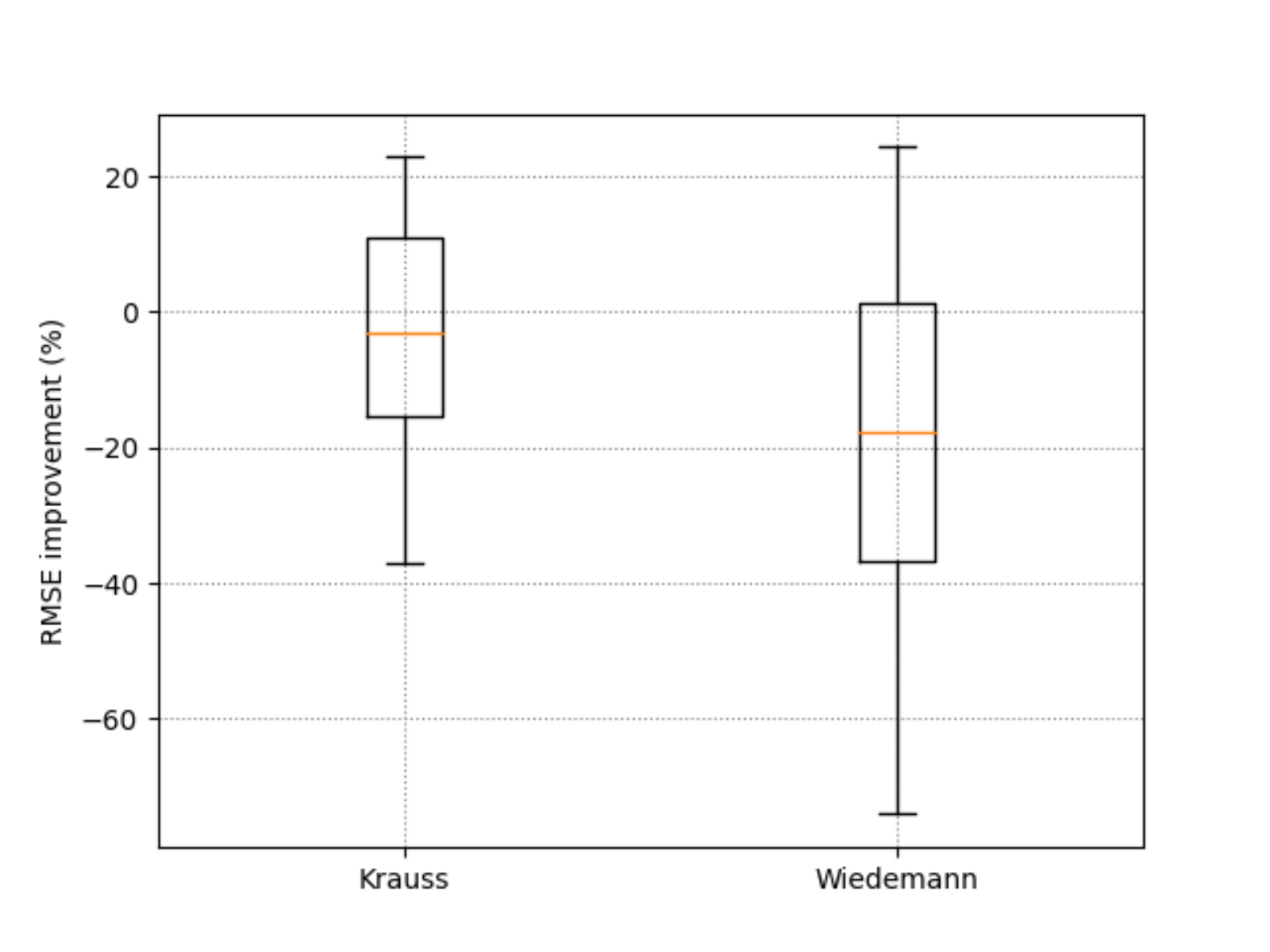}
\caption{{\it RMSE improvement} with {\it fixed parameters} for each following vehicle}
\label{figure:diff_one}
\end{figure}

We next evaluated the proposed method by applying it to each vehicle pair (94 pairs of leading and following vehicle data) for both the Krauss and Wiedemann models. The data of each vehicle pair were used to train and test, the same as in the previous experiment for the Krauss model. Figure \ref{figure:diff_one} shows each following vehicle’s {\it RMSE improvement} with {\it fixed parameters} in the Krauss and Wiedemann models. Note that we eliminated the top and bottom 10\% for the value of {\it RMSE improvement} as outlier.

\subsubsection{Online adaptation for clusters}
\begin{figure}[t!]
\centering
\includegraphics[width=80mm,bb=0 0 461 346]{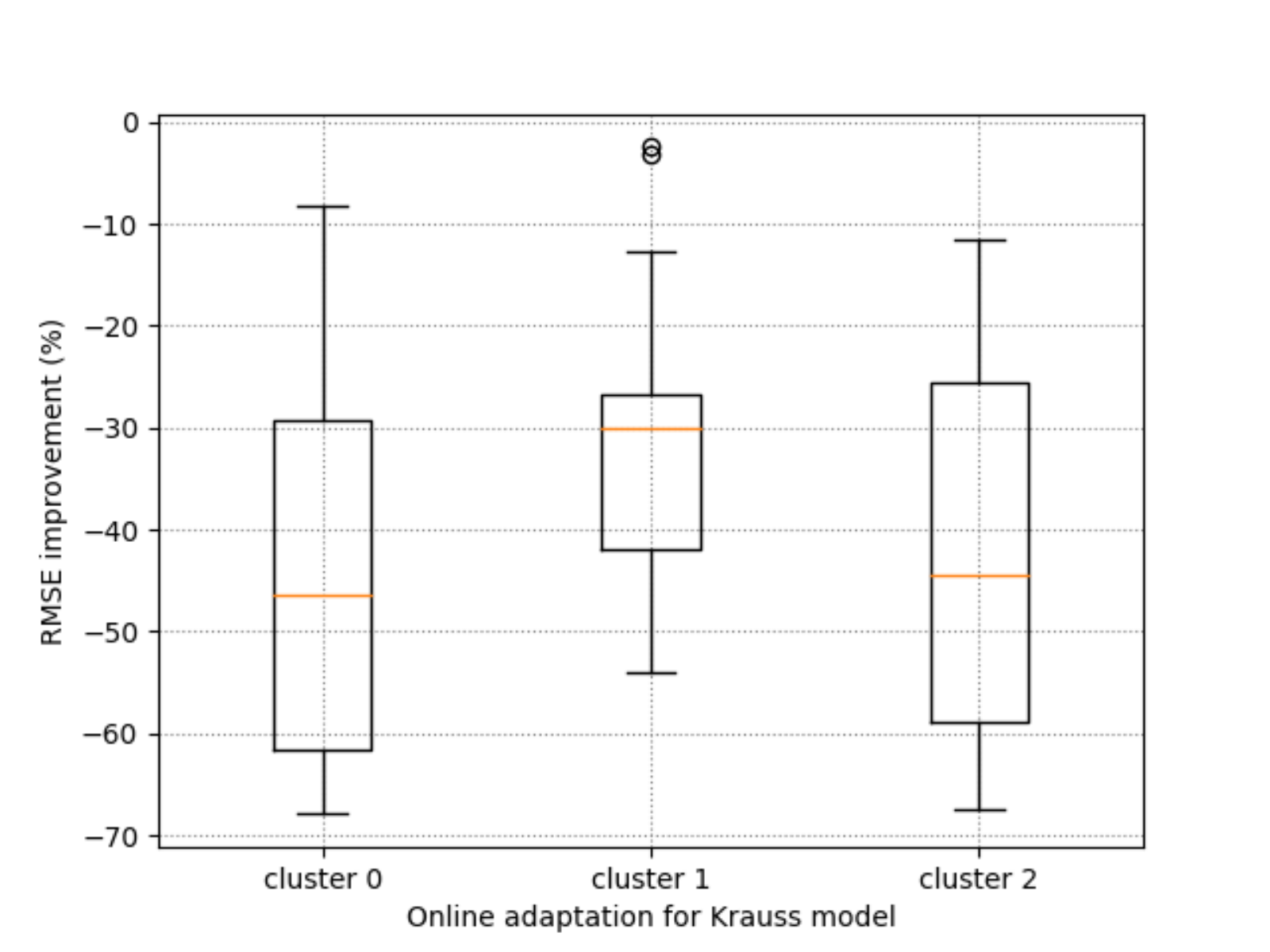}
\caption{{\it RMSE improvement} with {\it fixed parameters} for clusters (Krauss model)}
\label{figure:diff_cluster_kr}
\end{figure}

\begin{figure}[t!]
\centering
\includegraphics[width=80mm,bb=0 0 461 346]{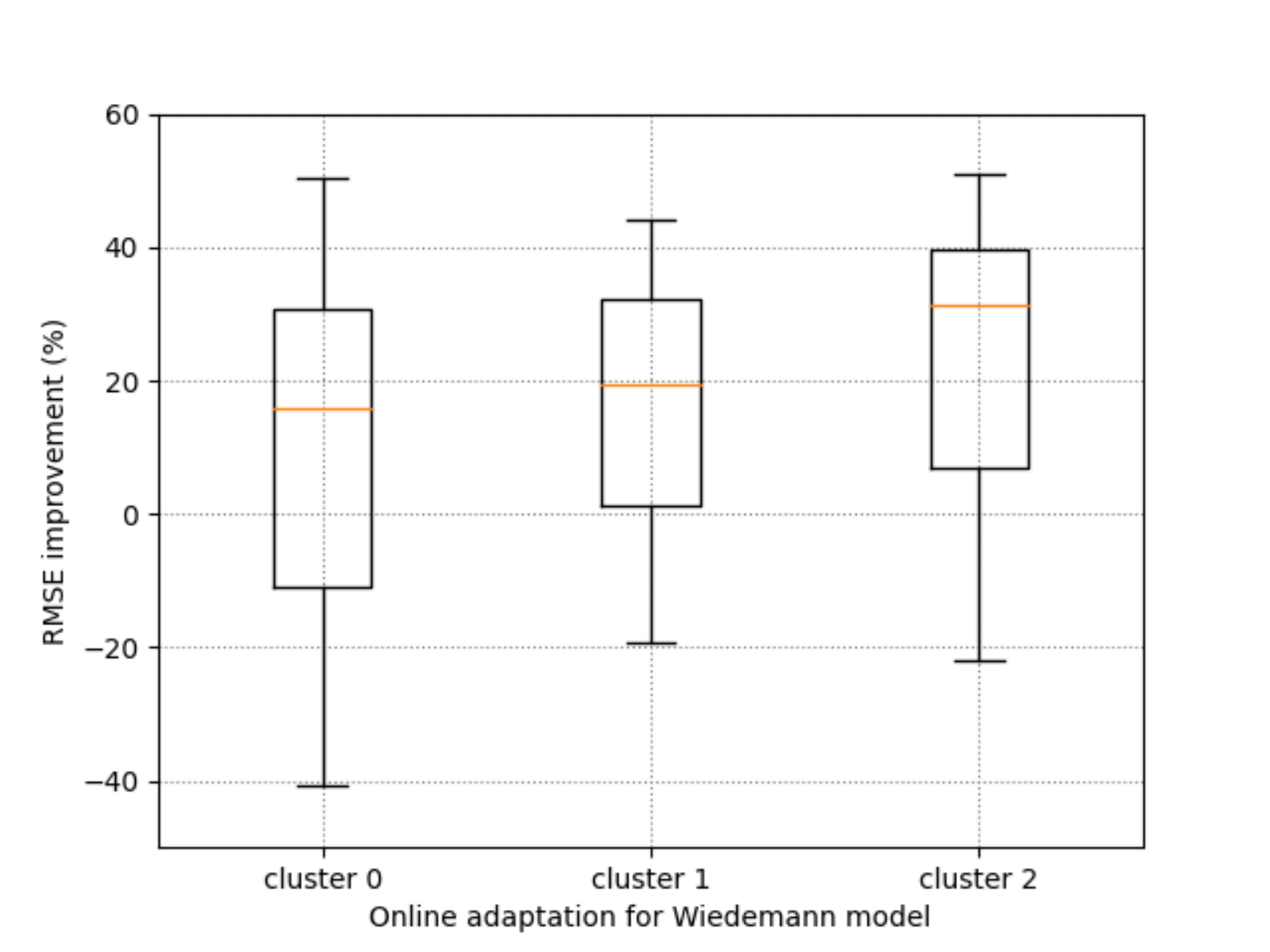}
\caption{{\it RMSE improvement} with {\it fixed parameters} for clusters (Wiedemann model)}
\label{figure:diff_cluster_w99}
\end{figure}

We then evaluated the {\it RMSE improvement} for each cluster that was divided into three by entropy weight. We used 5-fold cross validation for each cluster to train and test the vehicle data in a cluster. For example, in cluster-0, which had 24 vehicles, the data of 19 vehicles were used to train the GRU-based NN and the data of 5 vehicles were used  to test the NN. The training and testing were repeated for each 5-fold set. Figure \ref{figure:diff_cluster_kr} shows the {\it RMSE improvement}  with {\it fixed parameters} for clusters in the Krauss model, and Figure \ref{figure:diff_cluster_w99} shows the {\it RMSE improvement} with {\it fixed parameters} for clusters in the Wiedemann model.

\subsubsection{Demonstration of proposed method}
\begin{figure}[t!]
\centering
\includegraphics[width=60mm,bb=0 0 376 546]{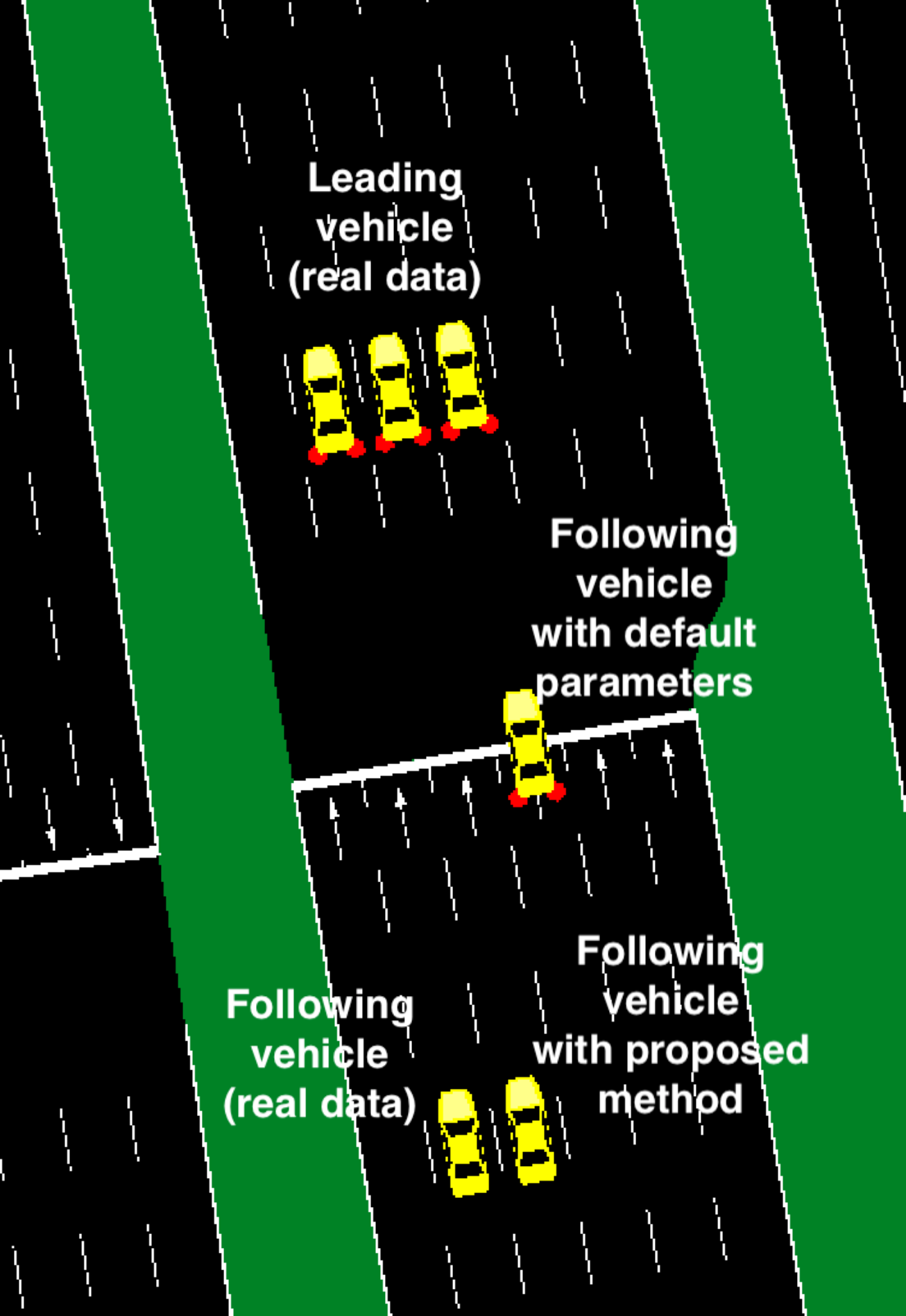}
\caption{Demonstration of proposed method on SUMO simulation}
\label{figure:sumo_replay}
\end{figure}

We demonstrated the proposed method on the microscopic traffic simulator SUMO. Figure \ref{figure:sumo_replay} shows a demonstration of three types of following vehicle simulation in SUMO; real vehicle in the dataset, calibrated data by the proposed method and {\it default parameters}. SUMO can import maps from OpenStreetMap \cite{OpenStreetMap}, and we imported the map of the I-80 targeted area. All the leading vehicles were replayed in accordance with the real data in the dataset, and following vehicles were simulated using real data, the proposed method, and {\it default parameters}. We confirmed that the proposed method is effective for a real simulator. 

\section{Discussion}
Regarding the challenge that a DM should simulate real human driving behavior, the proposed method enabled the calibration of a DM model (Wiedemann model) that is up to 26\% more accurate than {\it fixed parameters} for all clusters, as shown in Figure \ref{figure:diff_cluster_w99}. Average {\it RMSE improvement} for cluster-2 (aggressive driving) was greater than those for cluster-0 (conservative driving). This indicates that online adaptation of conservative driving is more difficult than aggressive driving, which may be due to conservative driving involving low velocity.

The {\it RMSE improvement}  between the proposed method and {\it fixed parameters} depended on the vehicle pair, as shown Figure \ref{figure:diff_one}. The average {\it RMSE improvement}  between both car-following models were less than zero, which means they were worse than {\it fixed parameters} for each following vehicle. However, the average {\it RMSE improvement}  for clusters were higher than zero. This is assumed due to the clusters having a greater amount of training data than each vehicle [pair?]. For example, the number of training data for vehicle ID=11 was only 134; however, cluster-0 had about 2500 time windows (=19 vehicles $\times$ 134 time windows).

The proposed method with the Wiedemann model achieved better accuracy than with the Krauss model, as shown in Figures \ref{figure:diff_cluster_kr} and \ref{figure:diff_cluster_w99}. This is considered due to  fact that the number of parameters for each model were different. The Krauss model has two parameters and Wiedemann model has ten parameters. 

\section{Conclusion and Future work}
The proposed method divided vehicle trajectory data in accordance with their driving styles and achieved more accurate than other DM-calibration methods. We applied the proposed method to two car-following models; however, we believe the method can also be applied to humanoid robots that imitate human behavior since such robots use models to simulate human actions. 

\bibliographystyle{ACM-Reference-Format}
\bibliography{drivingmodel}

\end{document}